%% file: main.tex
\documentclass[10pt,twocolumn,letterpaper]{article}

\PassOptionsToPackage{pdftex,unicode}{hyperref}

\def\CHK#1 {\textcolor{magenta}{{\bf [CHK:}~#1{\bf ]}}~}
\def\ADD#1 {\textcolor{cyan}{{\bf [To add:}~#1{\bf}]}~}

\newcommand{\ie}{\emph{i.e., }}

\usepackage[pagenumbers]{cvpr} 

\usepackage{graphicx}
\usepackage{amsmath}
\usepackage[accsupp]{axessibility}
\usepackage{amssymb}
\usepackage{booktabs}
\usepackage[noend]{algpseudocode}

\usepackage{algorithmicx,algorithm}

\usepackage{times}
\usepackage{pifont}       
\usepackage{bbding}       
\usepackage{setspace}
\usepackage{multirow}
\usepackage{cite}
\usepackage{comment}
\usepackage{booktabs}
\usepackage{arydshln}
\usepackage{color}
\usepackage{xcolor}
\usepackage{bm}
\usepackage[normalem]{ulem}
\usepackage{multirow}
\usepackage{cellspace}
\usepackage{colortbl}
\begin{document}

\title{StructAlign: Structured Cross-Modal Alignment for Continual Text-to-Video Retrieval}

\author{
Shaokun Wang,
Weili Guan,
Jizhou Han,
Jianlong Wu,
Yupeng Hu, 
Liqiang Nie\\ 
Harbin Institute of Technology (Shenzhen), Xi’an Jiaotong University, Shandong University\\
{\tt\small wangshaokun@hit.edu.cn}, {\tt\small honeyguan@gmail.com}, {\tt\small jizhou-han@stu.xjtu.edu.cn} \\
 {\tt\small wujianlong@hit.edu.cn}, {\tt\small huyupeng@sdu.edu.cn}, {\tt\small nieliqiang@gmail.com}}

\maketitle

\begin{abstract}
Continual Text-to-Video Retrieval (CTVR) is a challenging multimodal continual learning setting, where models must incrementally learn new semantic categories while maintaining accurate text-video alignment for previously learned ones, thus making it particularly prone to catastrophic forgetting. A key challenge in CTVR is feature drift, which manifests in two forms: intra-modal feature drift caused by continual learning within each modality, and non-cooperative feature drift across modalities that leads to modality misalignment. 
To mitigate these issues, we propose StructAlign, a structured cross-modal alignment method for CTVR. 
First, StructAlign introduces a simplex Equiangular Tight Frame (ETF) geometry as a unified geometric prior to mitigate modality misalignment. Building upon this geometric prior, we design a cross-modal ETF alignment loss that aligns text and video features with category-level ETF prototypes, encouraging the learned representations to form an approximate simplex ETF geometry. 
In addition, to suppress intra-modal feature drift, we design a Cross-modal Relation Preserving loss, which leverages complementary modalities to preserve cross-modal similarity relations, providing stable relational supervision for feature updates. 
By jointly addressing non-cooperative feature drift across modalities and intra-modal feature drift, StructAlign effectively alleviates catastrophic forgetting in CTVR. Extensive experiments on benchmark datasets demonstrate that our method shows competitive advantages over state-of-the-art continual retrieval approaches. 
\end{abstract}

\section{Introduction}
\begin{figure*}[t]
\centering
  	\includegraphics[width=1.0\linewidth]{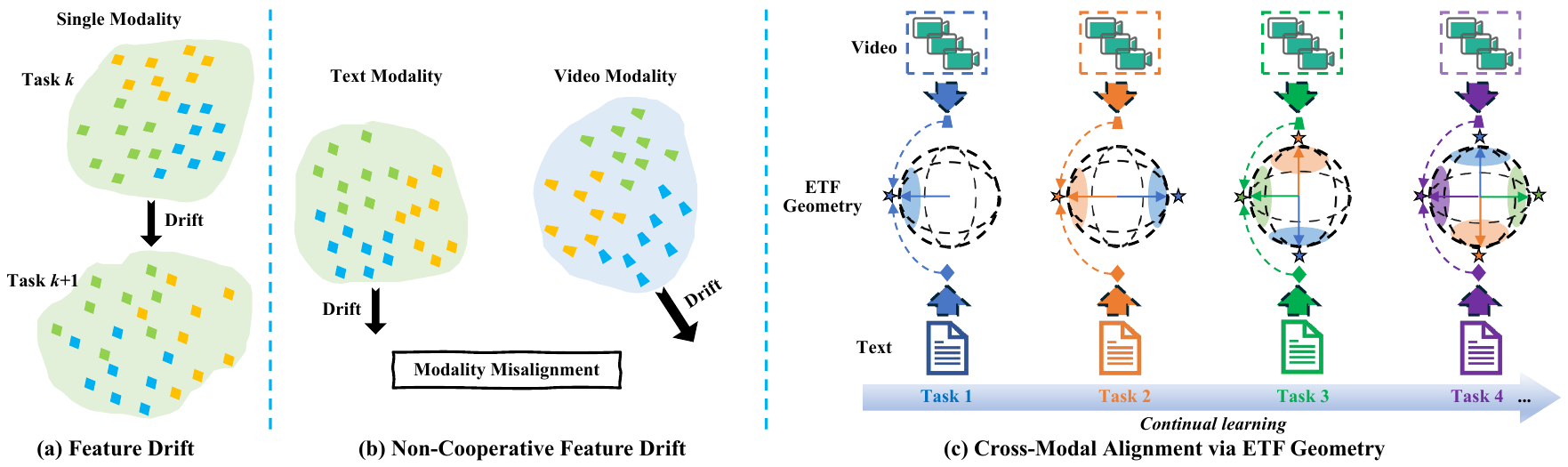}
  \caption
    {(a) In continual learning, feature drift refers to the phenomenon where feature representations shift toward new tasks, gradually deviating from feature representations learned for previous tasks. (b) In multimodal continual learning, features from different modalities tend to drift in a non-cooperative manner, which ultimately results in modality misalignment.  (c) Our StructAlign adopts a simplex ETF geometry as a unified geometric prior to mitigate modality misalignment during multimodal continual learning. 
  }
  \label{fig:concept}
\end{figure*}
Continual Learning (CL), also referred to as Lifelong or Incremental Learning, aims to enable models to continuously acquire new knowledge from a stream of tasks while retaining previously learned information. 
Early studies on CL have focused mainly on single-modality settings, such as image classification~\cite{dia} and object detection~\cite{LEA}.
Recently, CL has been extended to more complex and practically relevant multimodal tasks, including Image-Text Retrieval~\cite{cvlr}, Visual Question Answering~\cite{cvqa,vqacl}, and Text-to-Video Retrieval~\cite{ctvr}. 
In these Multimodal Continual Learning (MCL) settings, models are required to process multiple modalities with separate or weakly coupled encoders, while continuously adapting to new tasks.
This characteristic fundamentally differentiates MCL from its single-modality counterpart and makes it particularly vulnerable to catastrophic forgetting.

A primary cause of catastrophic forgetting is \emph{feature drift}, as illustrated in Fig.~\ref{fig:concept}(a), where strong data-fitting objectives drive feature representations to shift toward new tasks, gradually deviating from those learned for previous ones.
In MCL, feature drift manifests in two forms.
The first is \emph{intra-modal feature drift}, where features within each individual modality evolve as the model adapts to new tasks.
More critically, MCL introduces an additional challenge: \emph{modality misalignment}, which arises from \emph{non-cooperative feature drift} across modalities.
Since different modalities are typically encoded by independent encoders, gradients induced by new tasks update these encoders unevenly and asynchronously during continual learning.
As a result, features from different modalities may drift in different directions and with different magnitudes, gradually breaking the cross-modal alignment established on previous tasks.

This issue is particularly pronounced in text-to-video retrieval scenarios.
As shown in Fig.~\ref{fig:concept}(b), new tasks often contain novel visual content or temporal patterns, which strongly influence the video encoder and drive its feature space toward the new task distribution. 
In contrast, the text encoder usually receives weaker and less informative supervision, leading to relatively minor changes in its feature space. 
Such non-cooperative feature drift disrupts the shared feature space, causing semantically matched text and video features from previous tasks to become misaligned, and resulting in severe performance degradation. 

Motivated by these challenges, we study a novel and practically relevant MCL setting: 
\emph{Continual Text-to-Video Retrieval (CTVR)}, which was first introduced in~\cite{ctvr}. 
CTVR requires models to preserve fine-grained cross-modal alignment while continuously adapting to new categories. 
Consequently, it amplifies both intra-modal feature drift and non-cooperative feature drift across modalities, making catastrophic forgetting particularly severe in this setting.

In this paper, we propose StructAlign, a structured cross-modal alignment method for CTVR.  
To address non-cooperative feature drift across modalities, we introduce a simplex Equiangular Tight Frame~\cite{NC1} (ETF) geometry as a unified geometric prior. 
This prior organizes multimodal representations in a shared feature space, as shown in Fig.~\ref{fig:concept}(c). 
Based on this, we design a cross-modal ETF alignment loss that explicitly aligns text and video features with their corresponding ETF prototypes, encouraging the learned features to form an approximate simplex ETF geometry.  
By enforcing a well-separated category-level geometry and consistent cross-modal prototypes, this loss effectively reduces modality misalignment during continual learning. 
Even with improved cross-modal alignment, intra-modal feature drift remains a major source of catastrophic forgetting. 
To address this, we introduce a Cross-modal Relation Preserving (CRP) loss that leverages complementary modalities to preserve cross-modal similarity relations under the ETF-induced category geometry, thereby constraining intra-modal feature updates. 
Unlike conventional relational distillation methods designed for single-modality settings, CRP exploits cross-modal similarity relations as stable relational supervision, effectively regularizing intra-modal feature drift. 
By jointly addressing non-cooperative feature drift across modalities and intra-modal feature drift, our method effectively alleviates catastrophic forgetting in CTVR. 

Comprehensive experiments on benchmark datasets demonstrate the effectiveness of the proposed method. 
In-depth analyzes further show that the learned feature space approximates the ideal Simplex ETF geometry, yielding well-separated categories while maintaining concentrated yet diverse intra-category features.  
In summary, the main contributions are listed as follows:
\begin{itemize}
    \item We propose StructAlign\footnote{Code available at \url{https://github.com/Mysteriousplayer/SIGIR26-StructAlign}}, a structured cross-modal alignment framework for CTVR, which explicitly models and mitigates catastrophic forgetting induced by both intra-modal feature drift and non-cooperative feature drift across modalities. 
    \item
    We introduce a simplex ETF geometric prior together with a cross-modal ETF alignment loss to enforce a well-separated category-level structure in the shared embedding space. 
    In addition, we design a cross-modal relation preserving loss that leverages cross-modal similarity relations to constrain intra-modal feature updates during continual learning.
    \item
    Extensive experiments on benchmark datasets demonstrate that StructAlign achieves competitive performance compared to state-of-the-art continual retrieval methods. 
\end{itemize}

\section{Related Work}
\subsection{Continual Learning}
A range of methods has been proposed to address catastrophic forgetting in CL settings, which can be grouped into four categories: 1) Regularization-based approaches~\cite{LwF,BiCL,Balance,Tpcil,OSAKA} introduce additional loss terms to preserve previously acquired knowledge and  implicitly constrain feature drift;
2) Architecture-based approaches~\cite{AANet,DER,MGSVF,PackNet} dynamically expand or prune the network to reallocate parameters, accommodating new tasks while retaining old knowledge; 3) Rehearsal-based approaches~\cite{EndToEnd,AnchorReplay,CausalEffect,LargeScale,CuriosityDriven} store a small set of representative exemplars from prior data and replay them during continual learning to mitigate forgetting; 
and 4) Pre-trained model-based approaches~\cite{dia,l2p,DualPrompt,CoDa,EaSe,PromptIncremental,SPrompts,InstancePrompt} build upon large-scale pre-trained models and are often combined with Parameter-Efficient Tuning~\cite{pet,peng2025cia} (PET), achieving improvements in both forgetting resistance and overall performance. 
It should be noted that these four categories are not mutually exclusive. Many CL models~\cite{han2026goal,soul,polo,han_l,CrossSpace,Energy,wq1,wq2} adopt hybrid strategies that combine multiple approaches. 
Recently, increasing attention has been paid to MCL~\cite{mcl_1,mcl_2,mcl_3,cvqa,vqacl,cvlr,ctvr}, yet most existing methods still largely inherit techniques developed for unimodal continual learning.
However, MCL introduces challenges that go beyond a straightforward extension of unimodal CL, as complex interactions across modalities substantially exacerbate catastrophic forgetting~\cite{mcl_survey}.
In this paper, we propose StructAlign to mitigate modality misalignment caused by non-cooperative feature drift across modalities, a fundamental challenge in MCL. 

\subsection{Text-to-Video Retrieval}
Text-to-Video Retrieval (T2VR)~\cite{tvr23iccv,liu2025gaming, wang2024explicit,tvr_tmm22,finegrained,liu2018attentive} aims to retrieve video content that semantically matches a given textual query. 
Early approaches~\cite{tvr2022mm,tvr2021iccv,teachtext,t2vlad} rely on pre-trained modality-specific experts to extract offline features, followed by various strategies for multi-modal alignment, such as feature fusion modules~\cite{tvr2018eccv,tvr-mm22}, attention-based aggregation~\cite{tvr20eccv}, and cross-modal contrastive learning losses~\cite{TACo,ts2net}. 
With the advent of large-scale vision-language models like CLIP~\cite{clip}, recent methods~\cite{UMIVR,tvr_tmm23,xclip,centerclip,tvr2024mm,tvr23mm,uatvr,holistic,mpt} have demonstrated significantly improved performance, achieving more effective and generalizable text-video alignment. 
Early work, such as CLIP4Clip~\cite{CLIP4Clip}, pioneered the use of pre-trained CLIP models for text-to-video retrieval. 
Building on this paradigm, X-Pool~\cite{xpool} introduces a text-conditioned video pooling mechanism that leverages cross-modal attention to dynamically aggregate the most semantically relevant frames for a given textual query. 
CLIP-ViP~\cite{clipvip} enhances CLIP’s video understanding capability by incorporating auxiliary image captions and a proxy-guided video attention mechanism. 
PIG~\cite{pig} proposes a Hybrid-Tower CLIP-based architecture that combines the efficiency of two-tower models with the effectiveness of single-tower models. 
VIDEO-COLBERT~\cite{Video-ColBERT} advances fine-grained retrieval by enabling token-level interactions through MeanMaxSim over spatial and spatio-temporal levels. 
Despite their success, almost all existing methods assume access to the full training data in a single stage and yield models whose functionality remains fixed after training.  
In contrast, real-world scenarios involve a continuous stream of text-video pairs from previously unseen categories, resulting in evolving data distributions.  
These challenges highlight the importance of studying CTVR.

\subsection{Applications of Equiangular Tight Frame Geometry}
An Equiangular Tight Frame (ETF) characterizes a highly symmetric and well-separated geometric configuration, in which category prototypes are uniformly distributed in the embedding space with equal pairwise angles. 
This structure has been shown to naturally emerge in well-trained classifiers, a phenomenon commonly referred to as Neural Collapse (NC)~\cite{NC1}. 
Recent studies have extended NC-inspired formulations to various domains, including imbalanced classification~\cite{nc2}, federated learning~\cite{nc3}, incremental learning~\cite{nc4,nc6,nc5}, and generalized category discovery~\cite{hanconsistent}. 
In this work, we take a step further by introducing an ETF-based geometric constraint into the CTVR setting for the first time. 
Our goal is not to force features to collapse to a single point, as such collapse would harm retrieval tasks that require fine-grained and instance-level discrimination. 
Instead, we exploit the category-level structural property of the ETF as a geometric prior to organize the shared embedding space into well-separated semantic regions.
This design enables stable and consistent cross-modal alignment across tasks, while preserving sufficient intra-category variability to support effective text-to-video retrieval.

\begin{figure*}[t]
	\centering
	\includegraphics[width=1.0\textwidth]{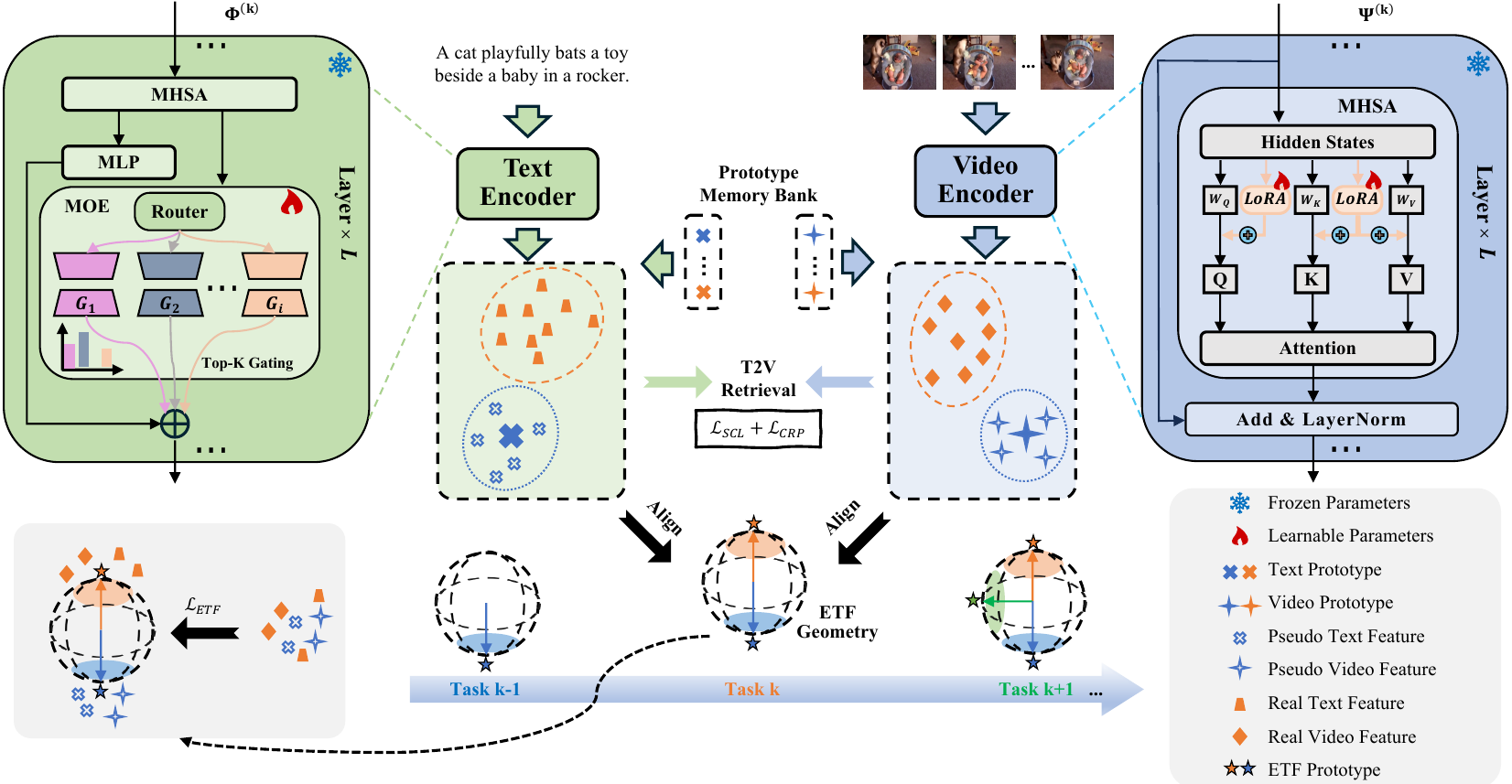} 
	\caption{
	The overview of StructAlign. Based on a simplex ETF geometric prior, StructAlign explicitly aligns both real features from the current task and pseudo features from previous tasks with their corresponding ETF prototypes via a cross-modal ETF alignment loss $\mathcal{L}_{ETF}$ (Sec.~4.2).  Moreover,  a cross-modal relation preserving loss $\mathcal{L}_{\mathrm{CRP}}$ is introduced to stabilize intra-modal feature updates by preserving cross-modal similarity relations under the ETF-induced geometry (Sec.~4.3 and Fig.~3).  
	}
	\label{figs:framework}
\end{figure*}
\section{Preliminary}
{\normalsize \bf Problem Setting.} 
In CTVR, the retrieval model learns query-video pairs on \(K\) training tasks, including a base task and \(K{-}1\) incremental tasks. 
At the \(k\)-th task, only the current dataset \(D^{k} = \{(\mathbf{t}_i, \mathbf{v}_i)\}_{i=1}^{|D^{k}|}\) is available, where \(\mathbf{t}_i\) and \(\mathbf{v}_i\) denote the \(i\)-th text query and video, respectively. 
Each video consists of \(M\) frames. 
Let \(\mathcal{Y}^{k}\) represent the set of new categories introduced in the task \(k\). 
These sets are disjoint across tasks, ensuring \(\forall i,j,\ \mathcal{Y}^{i} \cap \mathcal{Y}^{j} = \varnothing\). 
The model parameterized by \(\theta\) is sequentially trained on \(D^{1}, D^{2}, \dots, D^{K}\), and after learning from \(D^{k}\), it is evaluated on all encountered categories \(\mathcal{Y}^{1:k} = \mathcal{Y}^{1} \cup \mathcal{Y}^{2} \cup \cdots \cup \mathcal{Y}^{k}\). 
Given all test queries, the model retrieves the most relevant videos from all previously learned tasks. 

{\noindent\normalsize \bf Framework Overview.} Our framework builds upon frozen text and visual encoders, and incorporates lightweight parameter-efficient adaptation modules to support continual learning, together with a frame-word similarity function for text-to-video retrieval. 

Given a text query $\mathbf{t}$, we employ the CLIP~\cite{clip} \textbf{Text Encoder} $\Phi$ to extract word-level features:
\begin{equation}
\Phi(\mathbf{t}) = \{\mathbf{w}^n\}_{n=1}^{N},
\end{equation}
where $N$ denotes the length of the token sequence (the \texttt{[EOS]} token is omitted for brevity). 
To achieve a favorable plasticity-stability trade-off in continual learning, we freeze the parameters of the text encoder to preserve previously acquired knowledge, and inject a Mixture-of-Experts (MoE) module into the linear layers of each self-attention block to facilitate knowledge acquisition.
Specifically, the MoE module employs a router $\mathcal{R}(\cdot)$ to determine the contribution of each expert.
Given an input $x$, the gating weights
$\mathbf{G} = \{G_i\}_{i=1}^{I}$ are computed as:
\begin{equation}
\mathbf{G} = \mathrm{Softmax}\!\left( \mathrm{Top}\text{-}k_e \!\left( \mathcal{R}(x) \right) \right),
\end{equation}
where $\mathcal{R}(\cdot)$ projects $x$ to a one-dimensional vector indicating the activation likelihood of each expert.
The $\mathrm{Top}\text{-}k_e(\cdot)$ selects the $k_e$ most relevant experts while suppressing the remaining ones, and the $\mathrm{Softmax}(\cdot)$ function normalizes the selected weights to emphasize their relative contributions. 
Finally, the MoE output $x'$ is obtained as a weighted combination of expert outputs:
\begin{equation}
x' = \sum_{i=1}^{I} G_i \, E_i(x),
\end{equation}
where $E_i(\cdot)$ denotes the $i$-th expert LoRA.

Given a video $\mathbf{v}$ with $M$ frames, we employ the CLIP~\cite{clip} \textbf{Visual Encoder} $\Psi$ to extract frame-level features:
\begin{equation}
\Psi(\mathbf{v}) = \{\mathbf{f}^m\}_{m=1}^{M}.
\end{equation}
Analogous to the text encoder, we freeze all parameters of the visual encoder to preserve previously learned knowledge, and introduce LoRA modules to enable efficient task adaptation in a continual learning setting. 
Specifically, instead of modifying all the projection matrices in the self-attention mechanism, we inject LoRA modules into the query, key, and value projections of each self-attention block. Given an input token sequence \( x \) in a self-attention layer, the original projection matrices \( W_Q, W_K, W_V \in \mathbb{R}^{D \times D} \) are frozen. We augment the query, key, and value projections as follows: 
\begin{equation}
\begin{aligned}
\tilde{Q} &= x \left( W_Q + \Delta W_Q \right), \\
\tilde{K} &= x \left( W_K + \Delta W_K \right), \\
\tilde{V} &= x \left( W_V + \Delta W_V \right),
\end{aligned}
\end{equation}
where \( \Delta W_Q = A_Q B_Q \), \( \Delta W_K = A_K B_K \), and \( \Delta W_V = A_V B_V \). Here, \( A_\ast \in \mathbb{R}^{D \times r} \), \( B_\ast \in \mathbb{R}^{r \times D} \), and \( r \ll D \) represents the rank of the LoRA modules. 

{\noindent\normalsize \bf Frame-word Level Similarity.}
The frame-word level similarity measures the semantic alignment between a text sample $\mathbf{t}_i$ and its corresponding video $\mathbf{v}_j$ by aggregating fine-grained similarities between individual word and frame features:
\begin{equation}
\begin{aligned}
\mathrm{sim}(\mathbf{t}_i, \mathbf{v}_j) = \frac{1}{2} \Bigg(
& \frac{1}{N} \sum_{n=1}^{N} \max_{1 \le m \le M} 
\langle \mathbf{w}_i^n, \mathbf{f}_j^m \rangle \\
+\, 
& \frac{1}{M} \sum_{m=1}^{M} \max_{1 \le n \le N} 
\langle \mathbf{w}_i^n, \mathbf{f}_j^m \rangle
\Bigg),
\end{aligned}
\label{eq:frame-word-sim}
\end{equation}
where $N$ and $M$ are the numbers of words and frames in the pair, respectively. 
$\mathbf{w}_i^n \in \mathbb{R}^d$ and $\mathbf{f}_j^m \in \mathbb{R}^d$ represent the features of the $n$-th word and the $m$-th frame, and $\langle \mathbf{w}_i^n, \mathbf{f}_j^m \rangle$ denotes their cosine similarity. 
The first summation averages the maximum similarity that each word achieves with any frame, while the second summation performs the inverse operation over frames. 
The two terms are symmetrically averaged to ensure bidirectional consistency between text and video modalities.

\section{Methodology}
We first formulate an approximate Simplex ETF geometry to impose a category-level structure, and then introduce cross-modal ETF alignment and cross-modal relation preserving losses to address non-cooperative feature drift across modalities and intra-modal feature drift, respectively. 
Fig.~\ref{figs:framework} shows an overview of our method. 
\subsection{Approximate Simplex ETF Geometry for CTVR}

We introduce a simplex ETF structure as a geometric prior to organize the shared feature space for CTVR. 
The simplex ETF is initialized at the beginning of training and remains fixed throughout the training process. 
Instead of enforcing rigid feature collapse, it promotes a well-separated category-level geometry that supports cross-modal alignment while preserving intra-category variability. 
The term \emph{approximate} emphasizes that the ETF is employed as a soft geometric prior, rather than a strict convergence target. 

\noindent\textbf{Definition of a simplex ETF Structure.}  
The joint feature space is organized around a set of $C$ semantic prototypes, each corresponding to a distinct visual-textual concept.
We denote these prototypes by $P = [p_1, p_2, \dots, p_C] \in \mathbb{R}^{d \times C}$, which are encouraged to form a simplex ETF configuration.
In particular, an ideal reference ETF can be written as:
\begin{equation}
P^{\star} = \sqrt{\frac{C}{C - 1}} \, U \left(I_C - \frac{1}{C} \mathbf{1}_C \mathbf{1}_C^\top \right),
\label{eq:ETF-textvideo}
\end{equation}
where $U \in \mathbb{R}^{d \times C}$ is an orthogonal matrix satisfying $U^\top U = I_C$,
$I_C$ is the $C \times C$ identity matrix, and $\mathbf{1}_C$ is a $C$-dimensional all-ones vector. We assume $d \ge C$, which holds in practice. 
The prototypes are $\ell_2$-normalized and regularized to remain close to this reference geometry. 
Their pairwise inner products follow the simplex ETF structure:
\begin{equation}
p_c^\top p_{c'} =
\begin{cases}
1, & c=c',\\[4pt]
-\dfrac{1}{C-1}, & c \ne c'.
\end{cases}
\label{eq:approx-simplex}
\end{equation}
The off-diagonal inner product $-\dfrac{1}{C-1}$ characterizes the maximal equiangular configuration for $C$ prototypes.

\noindent\textbf{ETF-Induced Geometric Properties.}  
With ETF-based organization and symmetric cross-modal contrastive learning (temperature $\tau$), the joint feature space shows geometric properties that are beneficial for CTVR.  For simplicity, the modality index is omitted when it is not explicitly required.

\noindent\textit{(P1) Intra-category concentration.} 
Features belonging to the same semantic category are encouraged to concentrate around their corresponding prototype, without collapsing to a single point.
Let $\mu_{c,i}$ denote the feature of the $i$-th instance in category $c$, and $\mu_c$ the category mean.
The within-category covariance $\Sigma_W^{(c)} = \text{Avg}\{(\mu_{c,i} - \mu_c)(\mu_{c,i} - \mu_c)^\top\}$ is constrained to remain bounded: $\max_c \lambda_{\max}(\Sigma_W^{(c)}) \le \eta$, where $\eta = O(e^{-1/\tau})$ controls the allowable intra-category variation.

\noindent\textit{(P2) Inter-category equiangular separation.}  After centering the category means by the global mean $\mu_G = \text{Avg}_{c,i}(\mu_{c,i})$, the normalized category means $\hat{\mu}_c = \frac{\mu_c - \mu_G}{\|\mu_c - \mu_G\|}$ are encouraged to align with the ETF prototypes. 
As a result, their pairwise inner products approximately follow the simplex ETF structure:
\begin{equation}
\hat{\mu}_c^\top \hat{\mu}_{c'} =
\begin{cases}
1, & c=c',\\[4pt]
-\dfrac{1}{C-1} + \varepsilon_{c,c'}, & c \ne c',
\end{cases}
\quad \text{with } |\varepsilon_{c,c'}| \le \epsilon.
\end{equation}
Here, $\varepsilon_{c, c'}$ quantifies the deviation from the ideal equiangular configuration, and $\epsilon = O(\sqrt{\eta} + e^{-1/\tau})$.

\noindent\textit{(P3) Cross-modal prototype consistency.}  
For each category $c$, the text and video features are aligned with the same ETF prototype $p_c$.
Let $\hat{\mu}_c^t$ and $\hat{\mu}_c^v$ denote the normalized text and video category means, respectively.
Their cross-modal discrepancy is quantified by $\delta_c = 1 - \langle \hat{\mu}_c^t, \hat{\mu}_c^v \rangle$, with $\max_c \delta_c \le \gamma$, where $\gamma = O(e^{-1/\tau})$, and a smaller $\gamma$ indicates a stronger cross-modal geometric alignment.

\noindent\textbf{Discussion.}  
The bounded intra-category variation $\eta$, inter-category angular deviation $\epsilon$, and cross-modal discrepancy $\gamma$ provide a unified characterization of the approximate Simplex ETF geometry.
While the ETF prior specifies the desired category-level structure, symmetric cross-modal contrastive learning enforces these properties with controllable approximation errors. 

\subsection{Cross-modal ETF Alignment Loss}
To realize the approximate Simplex ETF geometry, we design a cross-modal ETF alignment loss that explicitly enforces the alignment of text and video features with their corresponding ETF prototypes. 

\noindent\textbf{Base Task.}
Given a text-video pair $(\mathbf{t}, \mathbf{v})$ with category label $c \in [1, C]$,
let $\Phi(\mathbf{t}) = \{\mathbf{w}^n\}_{n=1}^{N}$ and $\Psi(\mathbf{v}) = \{\mathbf{f}^m\}_{m=1}^{M}$ denote token-level text features and frame-level video features.
We first project token- and frame-level features into the shared ETF prototype space using modality-specific MLPs:
\begin{equation}
\tilde{\mathbf{w}}^n = g_t(\mathbf{w}^n), \qquad
\tilde{\mathbf{f}}^m = g_v(\mathbf{f}^m),
\label{eq:mlp-projection}
\end{equation}
where $g_t(\cdot)$ and $g_v(\cdot)$ map features to the same latent space as the ETF prototypes. 
Prototype-guided attention weights are computed as:
\begin{equation}
\alpha_n =
\frac{\exp\!\left( \langle \tilde{\mathbf{w}}^n, {p}_c \rangle \right)}
{\sum_{n'=1}^{N} \exp\!\left( \langle \tilde{\mathbf{w}}^{n'}, {p}_c \rangle \right)},
\quad
\beta_m =
\frac{\exp\!\left( \langle \tilde{\mathbf{f}}^m, {p}_c \rangle \right)}
{\sum_{m'=1}^{M} \exp\!\left( \langle \tilde{\mathbf{f}}^{m'}, {p}_c \rangle \right)},
\end{equation}
where ${p}_c$ denotes the ETF prototype for category $c$.
Modality-level representations are obtained via weighted pooling:
\begin{equation}
\bar{\mathbf{w}} = \sum_{n=1}^{N} \alpha_n \tilde{\mathbf{w}}^n,
\qquad
\bar{\mathbf{f}} = \sum_{m=1}^{M} \beta_m \tilde{\mathbf{f}}^m.
\end{equation}
The cross-modal ETF alignment loss for a mini-batch $\mathcal{B}$ is:
\begin{equation}
\mathcal{L}_{ETF}
=
\frac{1}{|\mathcal{B}|}
\sum_{(\bar{\mathbf{w}}, \bar{\mathbf{f}}, c) \in \mathcal{B}}
\Big[
(1 - \langle \hat{\mathbf{w}}, {p}_{c} \rangle)
+
(1 - \langle \hat{\mathbf{f}}, {p}_{c} \rangle)
\Big],
\label{eq:etf-alignment-batch}
\end{equation}
where $\hat{\mathbf{w}} = \bar{\mathbf{w}} / \|\bar{\mathbf{w}}\|$ and $\hat{\mathbf{f}} = \bar{\mathbf{f}} / \|\bar{\mathbf{f}}\|$.

\noindent\textbf{Incremental Task.}
In incremental tasks, text-video pairs from previously learned categories are no longer available.
To preserve embedding geometry and mitigate catastrophic forgetting, we leverage the learned modality-specific category prototypes (\ie ~category means) as anchors. 
Let $\hat{\mu}_c^t$ and $\hat{\mu}_c^v$ denote the normalized text and video category means.
For old categories, we synthesize pseudo features by adding Gaussian noise:
\begin{equation}
\tilde{\mathbf{w}}_c = \hat{\mu}_c^t + \boldsymbol{\epsilon}_c^t,
\qquad
\tilde{\mathbf{f}}_c = \hat{\mu}_c^v + \boldsymbol{\epsilon}_c^v,
\end{equation}
where $\boldsymbol{\epsilon}_c^t, \boldsymbol{\epsilon}_c^v \sim \mathcal{N}(\mathbf{0}, \sigma^2)$.
These pseudo features are defined in the same feature space after $g_t$ and $g_v$, ensuring geometric consistency with real samples. 

Modality-level representations for current task samples are obtained in the same way as in the base task.  
We then enforce prototype alignment for both real and pseudo features using Eq.~(\ref{eq:etf-alignment-batch}).  
This alignment ensures a consistent geometric structure across incremental tasks and alleviates catastrophic forgetting without requiring access to old data.

\subsection{Cross-modal Relation Preserving Loss}
We have designed a cross-modal ETF alignment loss to alleviate non-cooperative feature drift across modalities by enforcing category-level geometric consistency.
However, even with improved cross-modal alignment, intra-modal feature drift within each individual modality may still occur, which remains a major contributor to catastrophic forgetting. 
A direct solution is to constrain the features within each modality. 
However, existing approaches such as logit distillation~\cite{LwF} and relational distillation~\cite{LUCIR} are primarily designed for single-modality settings or typically impose rigid instance-level constraints.
Such constraints often overly restrict model plasticity, making them less suitable for CTVR.  

To address intra-modal feature drift in a more semantically grounded manner, we propose a Cross-modal Relation Preserving (CRP) loss. 
Instead of directly constraining individual features, CRP leverages the complementary modality to construct cross-modal relations. 
The key insight is that, while individual features may drift significantly during continual learning, the relative similarity relations across modalities are more stable.
In particular, these relations are partially preserved by the ETF-based geometric prior and can therefore serve as effective anchors to regularize intra-modal feature updates. 

Concretely, given a mini-batch of $\mathcal{B}$ text-video pairs $\{(\mathbf{t}_i, \mathbf{v}_i)\}_{i=1}^{\mathcal{B}}$, we compute the cross-modal similarity matrices $\mathbf{S}^{k}$ and $\mathbf{S}^{k-1}$ under the current model $\theta_k$ and the previous model $\theta_{k-1}$, respectively.
For each text sample $\mathbf{t}_i$, its similarity to all video samples $\{\mathbf{v}_j\}_{j=1}^{\mathcal{B}}$ under the current model is defined as:
\begin{equation}
\mathbf{S}_{i,j}^{k} = sim(\mathbf{t}_i, \mathbf{v}_j \,;\, \theta_k),
\end{equation}
and analogously under the previous model as:
\begin{equation}
\mathbf{S}_{i,j}^{k-1} = sim(\mathbf{t}_i, \mathbf{v}_j \,;\, \theta_{k-1}),
\end{equation}
where $\mathrm{sim}(\cdot,\cdot)$ is defined in Eq.~(\ref{eq:frame-word-sim}).

\begin{figure}[t]
\begin{center}
    \includegraphics[width=0.4\textwidth]{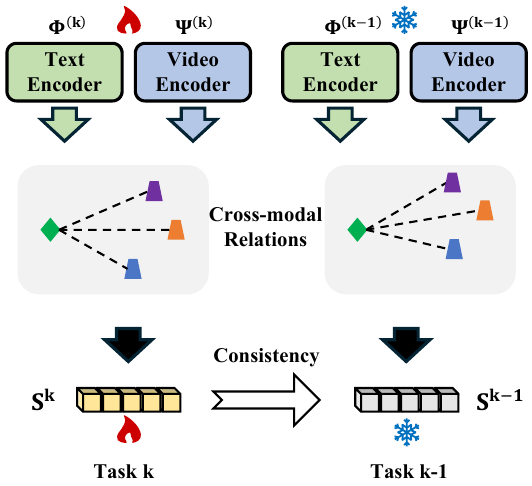}
\end{center}
  \caption{Illustration of the cross-modal relation preserving loss. By leveraging the complementary modality, cross-modal similarity matrices $\mathbf{S}^{k}$ and $\mathbf{S}^{k-1}$ are constructed under the current model $\theta_k$ and the previous model $\theta_{k-1}$, respectively. Treating $\mathbf{S}^{k-1}$ as a fixed relational anchor, the current similarity matrix $\mathbf{S}^{k}$ is optimized to preserve its relational structure. For clarity, the text modality is shown as the auxiliary modality, while the formulation is symmetric across modalities.}
  \label{figs:loss} 
\end{figure}

By treating $\mathbf{S}^{k-1}$ as a fixed relational anchor and encouraging $\mathbf{S}^{k}$ to preserve its structure, as shown in Fig.~\ref{figs:loss}, the CRP loss imposes a relation-level constraint on feature updates.  
Formally, it minimizes the Kullback-Leibler divergence between the softened similarity distributions, computed in a row-wise manner and averaged over the mini-batch: 
\begin{equation}
\mathcal{L}_{\mathrm{CRP}} 
=
\frac{1}{\mathcal{B}}
\sum_{i=1}^{\mathcal{B}}
\mathrm{KL}\!\left(
\mathrm{Softmax}\!\left(\mathbf{S}^{k-1}_{i,:} / \tau_2\right)
\;\middle\|\;
\mathrm{Softmax}\!\left(\mathbf{S}^{k}_{i,:} / \tau_2\right)
\right),
\label{crp}
\end{equation}
where $\mathbf{S}^{k}_{i,:}$ and $\mathbf{S}^{k-1}_{i,:}$ denote the $i$-th row of the similarity matrix under models $\theta_k$ and $\theta_{k-1}$, respectively. 
$\mathcal{B}$ is the mini-batch size, and $\tau_2$ is a temperature parameter.
\subsection{Optimization Objective}
We define a multi-part loss $\mathcal{L}$ to simultaneously facilitate new knowledge acquisition and old knowledge preservation in CTVR:
\begin{equation}
\mathcal{L} =
\begin{cases}
\mathcal{L}_{SCL} + \lambda_{1} \mathcal{L}_{ETF}, & k = 1,\\
\mathcal{L}_{SCL} + \lambda_{1} \mathcal{L}_{ETF} + \lambda_{2} \mathcal{L}_{CRP}, & k > 1,\\
\end{cases}
\end{equation}
where $\mathcal{L}_{ETF}$ and $\mathcal{L}_{CRP}$ are described in Eq.~(\ref{eq:etf-alignment-batch}) and Eq.~ (\ref{crp}), respectively. 
$\lambda_{1}$ and $\lambda_{2}$ indicate the hyper-parameters to balance the contributions of the two losses. 
The retrieval loss $\mathcal{L}_{SCL}$ is designed to pull semantically matched pairs closer in the shared feature space while pushing apart irrelevant ones.  
Specifically, $\mathcal{L}_{SCL}$ is formulated as a symmetric contrastive loss consisting of text-to-video and video-to-text components:
\begin{equation}
\mathcal{L}_{t2v} = -
\frac{1}{\mathcal{B}} 
\sum_{i=1}^{\mathcal{B}} 
\log 
\frac{
\exp\left( sim(\mathbf{t}_i, \mathbf{v}_i) / \tau \right)
}{
\sum_{j=1}^{\mathcal{B}} 
\exp\left( sim(\mathbf{t}_i, \mathbf{v}_j) / \tau \right)
},
\end{equation}
\begin{equation}
\mathcal{L}_{v2t} = -
\frac{1}{\mathcal{B}} 
\sum_{i=1}^{\mathcal{B}} 
\log 
\frac{
\exp\left( sim(\mathbf{t}_i, \mathbf{v}_i) / \tau \right)
}{
\sum_{j=1}^{\mathcal{B}} 
\exp\left( sim(\mathbf{t}_j, \mathbf{v}_i) / \tau \right)
},
\end{equation}
 \begin{equation}
\mathcal{L}_{SCL} = 1/2(\mathcal{L}_{v2t} + \mathcal{L}_{t2v}),
\end{equation}
where $|\mathcal{B}|$ denotes the mini-batch size,  $sim(\cdot,\cdot)$ represents the similarity function defined in Eq.~(\ref{eq:frame-word-sim}), and $\tau$ is a temperature parameter that controls the sharpness of the similarity distribution. 

\section{Experiments}
In the following part of this section, we first provide the experimental setup, then present the evaluation results, and finally report ablation studies, hyper-parameter studies, and in-depth analysis. 
The experiments are designed to answer the following questions: 
\begin{itemize}
    \item \textbf{RQ1:} How does StructAlign perform compared to state-of-the-art CTVR methods?
    \item \textbf{RQ2:} How do the proposed components in StructAlign contribute to overall performance?
    \item \textbf{RQ3:} How do hyper-parameter settings affect StructAlign performance, and which settings yield the best results?
    \item \textbf{RQ4:} What inter-category and intra-category geometric structures emerge in the learned feature space under the proposed method? 
\end{itemize}
\begin{table}[hb]
\centering
\caption{Datasets and continual learning protocol.}
\label{tab:protocol}
\scalebox{0.9}{
\begin{tabular}{lccc}
\toprule[1.5pt]
Dataset & \#Category & Shot/Category& \#Task \\
\midrule
MSRVTT       
& 20  
&  \multirow{2}{*}{16}  
& \multirow{2}{*}{$K{=}10,20$} \\

ACTNET  
& 200 
&  \\
\bottomrule[1.5pt]
\end{tabular}
}
\end{table}
\definecolor{oursgray}{gray}{0.93} 

\begin{table*}[t]
\centering
\caption{Comparison of method performance for CTVR on MSRVTT and ACTNET datasets with 10 and 20 tasks.
The top two methods are highlighted in \textbf{bold} and \underline{underlined}. The second row of each method reports the standard deviation over three runs.}
\resizebox{\textwidth}{!}{
\renewcommand{\arraystretch}{1.2}
\setlength{\tabcolsep}{2pt}
\begin{tabular}{l | ccccc c | ccccc c | ccccc c | ccccc c }
\toprule[1.5pt]
\multirow{2}{*}{Method} 
& \multicolumn{6}{c|}{MSRVTT-10} 
& \multicolumn{6}{c|}{MSRVTT-20}
& \multicolumn{6}{c|}{ACTNET-10}
& \multicolumn{6}{c}{ACTNET-20} \\

\cmidrule(lr){2-7}\cmidrule(lr){8-13}\cmidrule(lr){14-19}\cmidrule(lr){20-25}
& R@1$\uparrow$ & R@5$\uparrow$ & R@10$\uparrow$ & Med$\downarrow$ & Mean$\downarrow$ & BWF$\downarrow$
& R@1$\uparrow$ & R@5$\uparrow$ & R@10$\uparrow$ & Med$\downarrow$ & Mean$\downarrow$ & BWF$\downarrow$
& R@1$\uparrow$ & R@5$\uparrow$ & R@10$\uparrow$ & Med$\downarrow$ & Mean$\downarrow$ & BWF$\downarrow$
& R@1$\uparrow$ & R@5$\uparrow$ & R@10$\uparrow$ & Med$\downarrow$ & Mean$\downarrow$ & BWF$\downarrow$ \\
\midrule
\multirow{2}{*}{ZS CLIP}
& 22.14 & 41.24 & 51.34 & 10.00 & 117.48 & 0.00
& 22.14 & 41.24 & 51.34 & 10.00 & 117.48 & 0.00
& 14.89 & 34.97 & 47.78 & 12.00 & 84.02 & 0.00
& 14.89 & 34.97 & 47.78 & 12.00 & 84.02 & 0.00
 \\
& $\pm$0.00 & $\pm$0.00 & $\pm$0.00 & $\pm$0.00 & $\pm$0.00 & $\pm$0.00
& $\pm$0.00 & $\pm$0.00 & $\pm$0.00 & $\pm$0.00 & $\pm$0.00 & $\pm$0.00
& $\pm$0.00 & $\pm$0.00 & $\pm$0.00 & $\pm$0.00 & $\pm$0.00 & $\pm$0.00
& $\pm$0.00 & $\pm$0.00 & $\pm$0.00 & $\pm$0.00 & $\pm$0.00 & $\pm$0.00
 \\
\cline{2-25}
\multirow{2}{*}{CLIP4Clip}
& 23.57 & 44.76 & 54.48 & 8.00 & 80.23 & 0.61
& 21.79 & 42.13 & 52.52 & 9.00 & 86.07 & 1.02
& 17.85 & 41.05 & \underline{54.88} & \underline{8.67} & 54.97 & 0.75
& 17.07 & 39.96 & 53.43 & 9.47 & 47.38 & 0.45
\\
& $\pm$0.37 & $\pm$0.24 & $\pm$0.61 & $\pm$0.00 & $\pm$1.32 & $\pm$0.37
& $\pm$0.20 & $\pm$0.30 & $\pm$0.35 & $\pm$0.00 & $\pm$0.47 & $\pm$0.46
& $\pm$0.06 & $\pm$0.88 & $\pm$0.33 & $\pm$0.58 & $\pm$0.82 & $\pm$0.08
& $\pm$0.09 & $\pm$0.57 & $\pm$0.45 & $\pm$0.16 & $\pm$0.41 & $\pm$0.22
 \\
\cline{2-25}
\multirow{2}{*}{X-Pool}
& 19.60 & 39.80 & 49.49 & 11.00 & 94.89 & 0.28
& 15.98 & 34.21 & 44.34 & 15.33 & 105.97 & 1.37
& 17.99 & 39.81 & 52.38 & 9.67 & 60.49 & 0.37
& 16.57 & 39.83 & 51.82 & 10.22 & 55.00 & 0.31
 \\
& $\pm$0.35 & $\pm$0.63 & $\pm$0.54 & $\pm$0.00 & $\pm$1.70 & $\pm$0.40
& $\pm$0.75 & $\pm$1.52 & $\pm$0.87 & $\pm$0.58 & $\pm$1.97 & $\pm$0.23
& $\pm$0.54 & $\pm$0.87 & $\pm$1.06 & $\pm$0.58 & $\pm$4.98 & $\pm$0.39
& $\pm$0.38 & $\pm$0.47 & $\pm$0.35 & $\pm$0.19 & $\pm$1.58 & $\pm$0.43
 \\
\cline{2-25}
\multirow{2}{*}{CLIP-ViP}
& 21.56 & 44.19 & 53.43 & 8.00 & 86.71 & 0.49
& 19.74 & 41.25 & 50.61 & 10.00 & 93.95 & 0.73
& 17.01 & 38.73 & 52.01 & 9.67 & 59.66 & 0.56
& 16.02 & 37.29 & 50.92 & 10.58 & 48.78 & 0.73
\\
& $\pm$1.07 & $\pm$0.31 & $\pm$0.52 & $\pm$0.00 & $\pm$0.71 & $\pm$0.74
& $\pm$0.19 & $\pm$0.29 & $\pm$0.20 & $\pm$0.00 & $\pm$1.19 & $\pm$0.31
& $\pm$0.53 & $\pm$0.82 & $\pm$0.75 & $\pm$0.58 & $\pm$1.50 & $\pm$0.21
& $\pm$0.19 & $\pm$0.21 & $\pm$0.71 & $\pm$0.38 & $\pm$1.16 & $\pm$0.13
 \\
\cline{2-25}
\multirow{2}{*}{LwF}
& 23.85 & 45.30 & 55.68 & 7.33 & 76.46 & 1.68
& 22.06 & 42.77 & 52.69 & 9.00 & 85.27 & 1.65
& 17.56 & 40.18 & 53.67 & 9.00 & 55.33 & 0.63
& 16.36 & \underline{40.14} & 53.29 & 9.44 & \underline{45.94} & 0.93
\\
& $\pm$0.09 & $\pm$0.26 & $\pm$0.32 & $\pm$0.58 & $\pm$0.44 & $\pm$0.59
& $\pm$0.44 & $\pm$1.33 & $\pm$0.91 & $\pm$1.00 & $\pm$3.99 & $\pm$0.72
& $\pm$0.12 & $\pm$0.20 & $\pm$0.41 & $\pm$0.00 & $\pm$1.86 & $\pm$0.42
& $\pm$0.31 & $\pm$0.31 & $\pm$0.46 & $\pm$0.25 & $\pm$2.22 & $\pm$0.14
\\
\cline{2-25}
\multirow{2}{*}{VR-LwF}
& 24.49 & 45.59 & 55.45 & 7.33 & 74.89 & 1.22
& 22.39 & 43.27 & 53.33 & 8.67 & 82.04 & 1.44
& 18.08 & \textbf{41.44} & \textbf{54.98} & \textbf{8.50} & \underline{53.28} & 0.68
& 17.21 & \textbf{40.96} & \textbf{54.18} & \textbf{9.00} & \textbf{44.45} & 0.58
 \\
& $\pm$0.20 & $\pm$1.14 & $\pm$0.89 & $\pm$0.58 & $\pm$2.56 & $\pm$0.46
& $\pm$0.43 & $\pm$0.43 & $\pm$0.96 & $\pm$0.58 & $\pm$2.16 & $\pm$0.16
& $\pm$0.55 & $\pm$0.36 & $\pm$0.45 & $\pm$0.50 & $\pm$2.39 & $\pm$0.47
& $\pm$0.36 & $\pm$0.26 & $\pm$0.15 & $\pm$0.00 & $\pm$0.70 & $\pm$0.13
 \\
\cline{2-25}
\multirow{2}{*}{ZSCL}
& 23.99 & 45.15 & 54.77 & 8.00 & 79.69 & 0.10
& 21.47 & 41.61 & 52.05 & 9.33 & 88.45 & 0.91
& 17.67 & \underline{41.05} & 54.05 & 9.00 & 55.74 & 0.35
& 16.83 & 38.90 & 52.07 & 9.33 & 65.03 & 0.70
 \\
& $\pm$0.44 & $\pm$0.33 & $\pm$0.24 & $\pm$0.00 & $\pm$0.95 & $\pm$0.78
& $\pm$0.77 & $\pm$0.87 & $\pm$0.92 & $\pm$0.58 & $\pm$6.38 & $\pm$0.33
& $\pm$0.55 & $\pm$0.47 & $\pm$0.15 & $\pm$0.00 & $\pm$0.19 & $\pm$0.45
& $\pm$0.17 & $\pm$0.55 & $\pm$0.31 & $\pm$0.58 & $\pm$1.59 & $\pm$0.08
 \\
\cline{2-25}
\multirow{2}{*}{MoE-Adapter}
& 22.92 & 42.76 & 52.11 & 9.00 & 105.70 & 0.14
& 22.70 & 41.96 & 51.82 & 9.00 & 112.86 & 0.01
& 16.63 & 37.29 & 50.36 & 10.33 & 70.49 & \underline{-0.15}
& 15.77 & 36.27 & 49.32 & 11.00 & 77.65 & \underline{-0.01}
\\
& $\pm$0.09 & $\pm$0.24 & $\pm$0.09 & $\pm$0.00 & $\pm$2.66 & $\pm$0.11
& $\pm$0.14 & $\pm$0.16 & $\pm$0.10 & $\pm$0.00 & $\pm$0.34 & $\pm$0.05
& $\pm$0.55 & $\pm$0.48 & $\pm$0.89 & $\pm$0.58 & $\pm$3.70 & $\pm$0.08
& $\pm$0.11 & $\pm$0.14 & $\pm$0.24 & $\pm$0.00 & $\pm$1.23 & $\pm$0.27
 \\
\cline{2-25}
\multirow{2}{*}{TVR+CL}
& 22.47 & 43.71 & 53.59 & 8.00 & 82.97 & 0.46
& 21.28 & 42.28 & 51.82 & 9.67 & 89.22 & 1.28
& 16.88 & 38.87 & 51.82 & 9.67 & 61.16 & 0.44
& 16.37 & 37.78 & 50.51 & 10.33 & 66.80 & 0.76
\\
& $\pm$0.55 & $\pm$0.42 & $\pm$0.28 & $\pm$0.00 & $\pm$1.31 & $\pm$0.17
& $\pm$0.66 & $\pm$1.10 & $\pm$0.63 & $\pm$0.58 & $\pm$3.44 & $\pm$0.73
& $\pm$0.62 & $\pm$0.70 & $\pm$1.45 & $\pm$0.58 & $\pm$4.34 & $\pm$0.27
& $\pm$0.27 & $\pm$1.08 & $\pm$1.11 & $\pm$0.58 & $\pm$5.46 & $\pm$0.26
 \\
\cline{2-25}
\multirow{2}{*}{StableFusion}
& \underline{25.87} & \underline{45.91} & \underline{56.03} & \underline{7.00} & \underline{74.70} & \underline{-0.45}
& \underline{25.16} & \underline{45.53} & \underline{55.10} & \underline{7.33} & \underline{77.79} & \underline{-0.70}
& \underline{18.21} & 40.45 & 53.94 & 9.00 & 56.14 & -0.01
& \underline{17.71} & 39.40 & 52.76 & \textbf{9.00} & 62.22 & 0.04
\\
& $\pm$0.33 & $\pm$0.03 & $\pm$0.64 & $\pm$0.00 & $\pm$1.06 & $\pm$0.22
& $\pm$0.14 & $\pm$0.42 & $\pm$0.07 & $\pm$0.58 & $\pm$0.57 & $\pm$0.25
& $\pm$0.33 & $\pm$0.29 & $\pm$0.34 & $\pm$0.00 & $\pm$0.63 & $\pm$0.40
& $\pm$0.25 & $\pm$0.08 & $\pm$0.15 & $\pm$0.00 & $\pm$1.99 & $\pm$0.06
 \\
\midrule
\multirow{2}{*}{StructAlign}
& \cellcolor{oursgray}\textbf{25.98}
& \cellcolor{oursgray}\textbf{46.51}
& \cellcolor{oursgray}\textbf{57.13}
& \cellcolor{oursgray}\textbf{6.67}
& \cellcolor{oursgray}\textbf{73.27}
& \cellcolor{oursgray}\textbf{-0.47}

& \cellcolor{oursgray}\textbf{25.37}
& \cellcolor{oursgray}\textbf{45.94}
& \cellcolor{oursgray}\textbf{56.01}
& \cellcolor{oursgray}\textbf{7.00}
& \cellcolor{oursgray}\textbf{76.63}
& \cellcolor{oursgray}\textbf{-0.85}

& \cellcolor{oursgray}\textbf{18.26}
& \cellcolor{oursgray}40.31
& \cellcolor{oursgray}54.68
& \cellcolor{oursgray}9.00
& \cellcolor{oursgray}\textbf{48.82}
& \cellcolor{oursgray}\textbf{-0.68}

& \cellcolor{oursgray}\textbf{17.89}
& \cellcolor{oursgray}39.69
& \cellcolor{oursgray}\underline{53.98}
& \cellcolor{oursgray}\textbf{9.00}
& \cellcolor{oursgray}52.19
& \cellcolor{oursgray}\textbf{-0.54}

\\

& \cellcolor{oursgray}$\pm$0.20
& \cellcolor{oursgray}$\pm$0.80
& \cellcolor{oursgray}$\pm$0.31
& \cellcolor{oursgray}$\pm$0.58
& \cellcolor{oursgray}$\pm$1.08
& \cellcolor{oursgray}$\pm$0.07

& \cellcolor{oursgray}$\pm$0.23
& \cellcolor{oursgray}$\pm$0.17
& \cellcolor{oursgray}$\pm$0.40
& \cellcolor{oursgray}$\pm$0.00
& \cellcolor{oursgray}$\pm$0.61
& \cellcolor{oursgray}$\pm$0.47

& \cellcolor{oursgray}$\pm$0.20
& \cellcolor{oursgray}$\pm$0.28
& \cellcolor{oursgray}$\pm$0.36
& \cellcolor{oursgray}$\pm$0.00
& \cellcolor{oursgray}$\pm$0.49
& \cellcolor{oursgray}$\pm$0.10

& \cellcolor{oursgray}$\pm$0.11
& \cellcolor{oursgray}$\pm$0.30
& \cellcolor{oursgray}$\pm$0.15
& \cellcolor{oursgray}$\pm$0.00
& \cellcolor{oursgray}$\pm$2.08
& \cellcolor{oursgray}$\pm$0.25
 \\

\midrule
\multirow{2}{*}{Upper Bound}
& 25.86 & 48.34 & 58.96 & 6.00 & 65.10 & -0.17
& 25.80 & 48.54 & 58.84 & 6.00 & 65.13 & 0.23
& 19.66 & 44.57 & 59.18 & 7.00 & 37.16 & 0.27
& 19.78 & 44.63 & 58.96 & 7.00 & 36.18 & 0.16
 \\
& $\pm$0.67 & $\pm$0.25 & $\pm$0.12 & $\pm$0.00 & $\pm$1.09 & $\pm$1.21
& $\pm$0.45 & $\pm$1.04 & $\pm$0.65 & $\pm$0.00 & $\pm$1.79 & $\pm$0.44
& $\pm$0.08 & $\pm$0.17 & $\pm$0.23 & $\pm$0.00 & $\pm$0.18 & $\pm$0.10
& $\pm$0.17 & $\pm$0.31 & $\pm$0.24 & $\pm$0.00 & $\pm$0.76 & $\pm$0.31
 \\
\bottomrule[1.5pt]
\end{tabular}
}
\label{talbe_m}
\end{table*}
\begin{table*}[]
\caption{
Ablation studies of StructAlign on MSRVTT and ACTNET, where the framework is described in Sec.3, ${CETF}$ denotes the cross-modal ETF alignment loss based on a simplex ETF structure, and  $\mathcal{L}_{CRP}$ denotes the cross-modal relation preserving loss. 
}
\centering
\scalebox{0.65}{
\begin{tabular}{l|cccc|cccc|cccc|cccc}
\toprule[1.5pt]
\textbf{Dataset}   
& \multicolumn{8}{c|}{MSRVTT} 
& \multicolumn{8}{c}{ACTNET} \\ 
\hline
\textbf{\#Task}   
& \multicolumn{4}{c}{10}  
& \multicolumn{4}{|c}{20} 
& \multicolumn{4}{|c}{10}  
& \multicolumn{4}{|c}{20} \\ 
\hline
\textbf{Metric}
& R@1$\uparrow$ & R@5$\uparrow$ & R@10$\uparrow$ & MeanR$\downarrow$ 
& R@1$\uparrow$ & R@5$\uparrow$ & R@10$\uparrow$ & MeanR$\downarrow$ 
& R@1$\uparrow$ & R@5$\uparrow$ & R@10$\uparrow$ & MeanR$\downarrow$ 
& R@1$\uparrow$ & R@5$\uparrow$ & R@10$\uparrow$ & MeanR$\downarrow$ \\
\hline
Framework 
& 22.71 & 42.61 & 52.81 & 100.97 
& 21.82 & 42.40 & 53.06 & 83.26
& 14.84 & 35.63 & 47.94 & 60.12 
& 15.71 & 36.47 & 50.50 & 59.75 \\
\hline
Framework + $\mathcal{L}_{CRP}$
& 24.65 & 45.28 & 55.69 & 77.28 
& 23.71 & 44.61 & 54.08 & 79.67
& 16.78 & 38.05 & 51.67 & 53.32 
& 17.07 & 38.66 & 52.78 & 54.08 \\
\hline
Framework + ${CETF}$
& 24.98 & 45.72 & 56.32 & 75.45 
& 24.62 & 45.12 & 55.33 & 79.51
& 17.16 & 39.11 & 53.78 & 51.68 
& 17.00 & 39.14 & 53.65 & 54.98 \\
\hline
\rowcolor[HTML]{EFEFEF} 
StructAlign  
& \textbf{25.98} & \textbf{46.51} & \textbf{57.13} & \textbf{73.27} 
& \textbf{25.37} & \textbf{45.94} & \textbf{56.01} & \textbf{76.63}
& \textbf{18.26} & \textbf{40.31} & \textbf{54.68} & \textbf{48.82} 
& \textbf{17.89} & \textbf{39.69} & \textbf{53.98} & \textbf{52.19} \\
\bottomrule[1.5pt]
\end{tabular}
}
\label{tbl2}
\end{table*}
\begin{figure}[]
\begin{center}
    \includegraphics[width=0.47\textwidth]{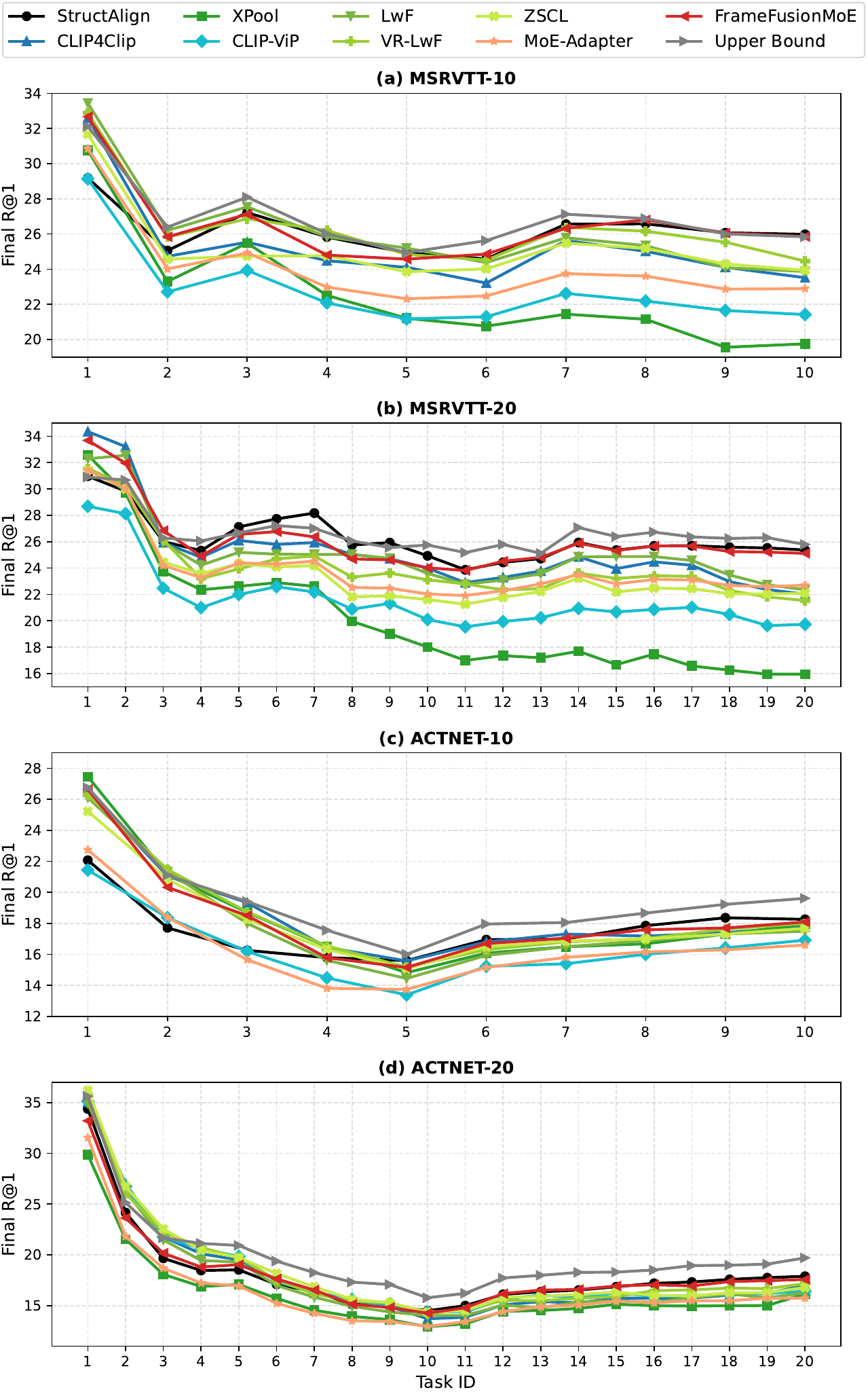}
\end{center}
  \caption{The comparison performance on MSRVTT (a)-(b) and ACTNET (c)-(d).}
  \label{figs:zxt} 
\end{figure}
\begin{figure}[]
\begin{center}
    \includegraphics[width=0.47\textwidth]{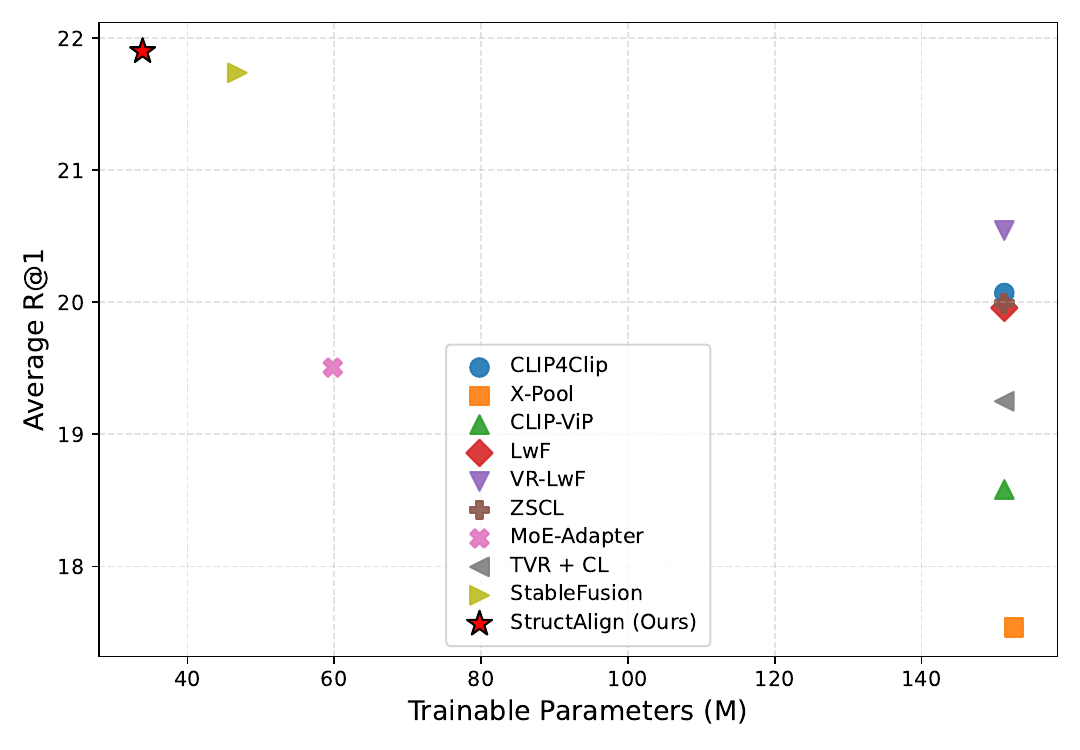}
\end{center}
  \caption{Trade-off between retrieval performance and trainable Parameters.}
  \label{figs:pvp} 
\end{figure}
\begin{figure*}[]
\centering
  	\includegraphics[width=1.0\linewidth]{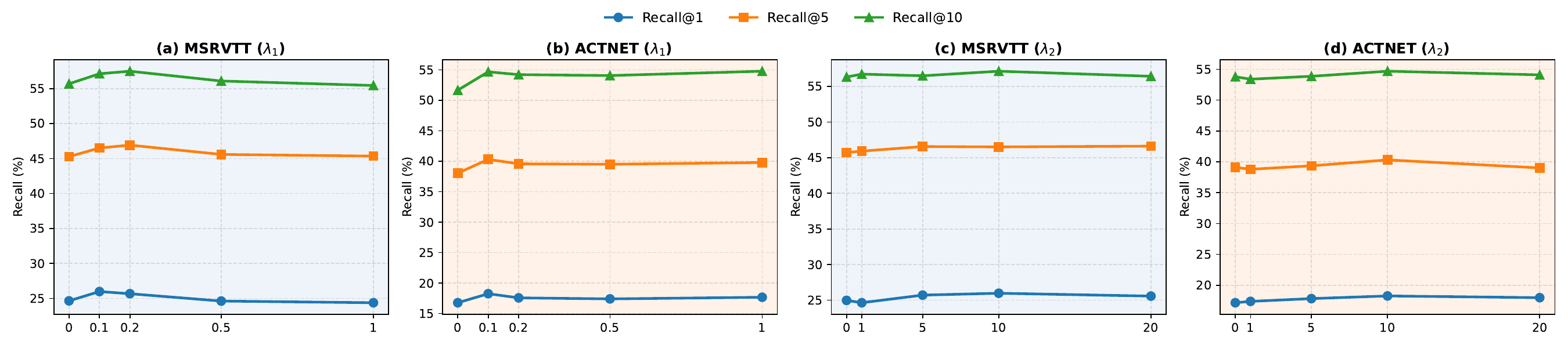}
  \caption
    {Hyper-parameter studies of $\lambda_{1}$ and $\lambda_{2}$ on MSRVTT and ACTNET.  
  }
  \label{fig:ab}
\end{figure*}
\begin{figure*}[]
\centering
  	\includegraphics[width=1.0\linewidth]{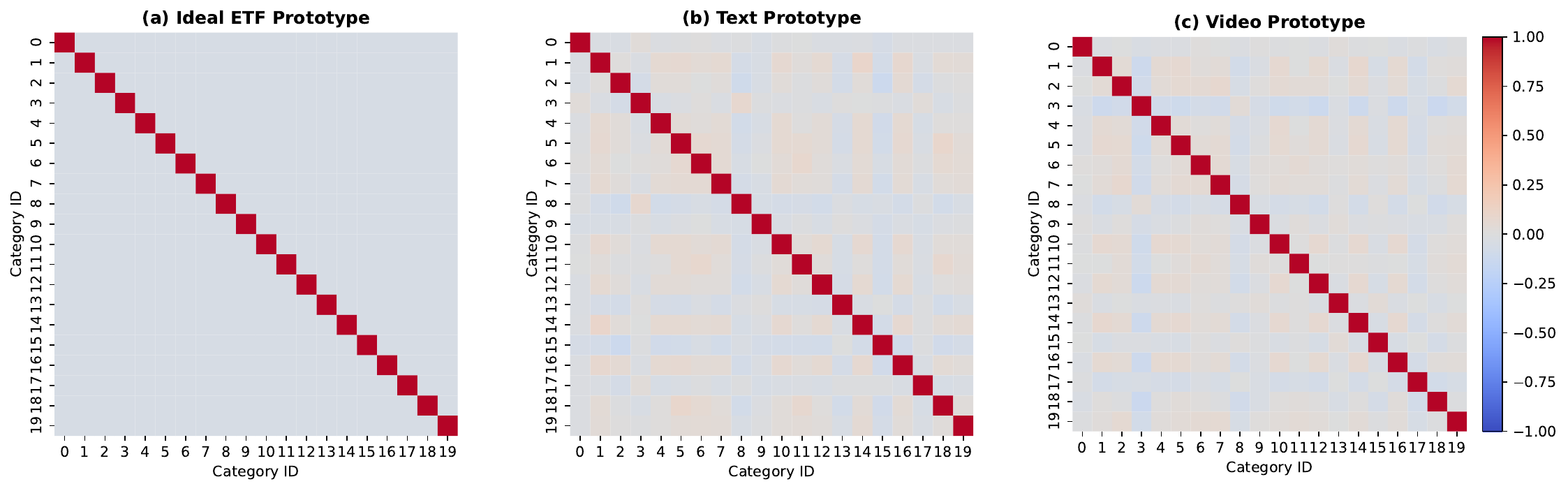}
  \caption
    {Pairwise cosine similarity matrices of (a) ideal Simplex ETF prototypes, (b) learned text prototypes, and (c) learned video prototypes.  
  }
  \label{fig:cf}
\end{figure*}
\begin{figure}[t]
\begin{center}
    \includegraphics[width=0.45\textwidth]{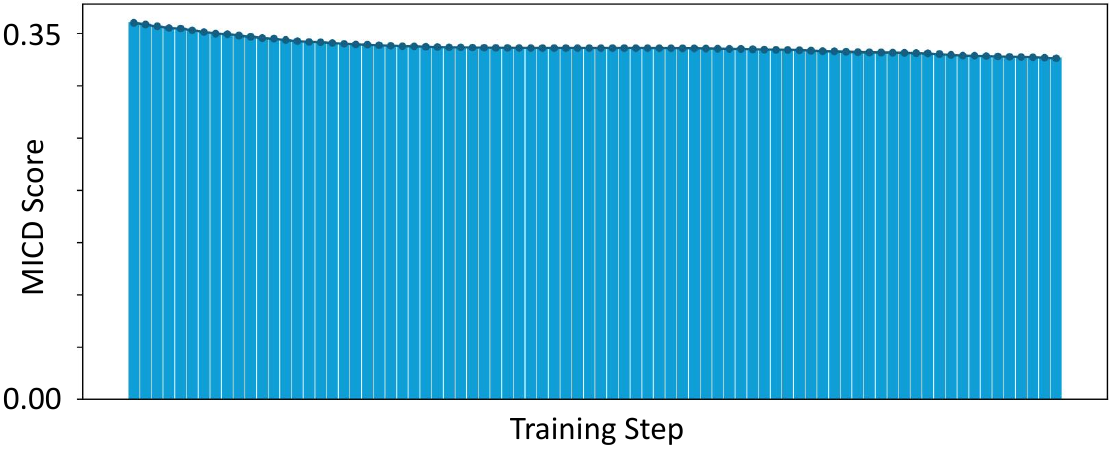}
\end{center}
  \caption{Trend of Mean Intra-Category Dispersion score over training steps.}
  \vspace{-2ex}
  \label{figs:micd} 
\end{figure}
\subsection{Experimental Setup}
{\normalsize \bf Datasets.}
We conduct extensive experiments on two widely used video-language benchmarks: \textbf{MSRVTT}~\cite{msrvtt} and \textbf{ACTNET}~\cite{activitynet}. \textbf{MSRVTT} contains 10,000 short video clips, each paired with approximately 20 human-annotated textual descriptions. 
\textbf{ACTNET} consists of around 20,000 long and untrimmed videos covering 200 human activity categories collected from YouTube. 
We follow the CTVR protocol proposed in~\cite{ctvr}, where all categories are evenly divided into \textit{K} tasks (with \textit{K} = 10 or 20), as shown in Table~\ref{tab:protocol}. 

{\noindent\normalsize \bf Implementation Details.}
The video and text encoders are initialized with the pre-trained CLIP ViT-B/32~\cite{clip}, and all pre-trained weights are frozen during training. StructAlign is implemented in PyTorch. It is trained on a single NVIDIA A100 Tensor Core GPU for 20 epochs with a batch size of 32. In the base task, the initial learning rate is set to $2\times10^{-5}$, while for incremental tasks, it is set to $3\times10^{-6}$ for MSRVTT and $6\times10^{-6}$ for ACTNET. All comparison methods use the same training samples. The hyper-parameters $\lambda_1$ and $\lambda_2$ are set to 0.1 and 10, respectively.

{\noindent\normalsize \bf Comparison Methods.} 
We compare our method with three categories of competitive baselines to comprehensively evaluate its effectiveness: 

\textbf{(1) State-of-the-art CTVR method}, \ie~StableFusion~\cite{ctvr}, which represents the current leading solution; 

\textbf{(2) Classical continual learning methods}, including LwF~\cite{LwF}, VR-LwF~\cite{vrlwf}, ZSCL~\cite{zscl}, and MoE-Adapter~\cite{moeada}, which are originally designed for single modality continual learning but can be adapted to the CTVR setting; 

\textbf{(3) Representative text-video retrieval methods}, including CLIP4Clip~\cite{CLIP4Clip}, X-Pool~\cite{xpool}, and CLIP-ViP~\cite{clipvip}, which reflect strong retrieval capabilities in standard non-incremental settings. 

In addition, we report the Upper Bound performance, where the model is trained using the complete dataset from all tasks simultaneously, serving as an oracle reference to measure the performance gap induced by the continual learning setting. 

{\noindent\normalsize \bf  Evaluation Metrics.}
We evaluate the proposed model using two categories of metrics. 
\textbf{(1) Retrieval Performance Metrics.} To assess retrieval capability, we report Recall@1, Recall@5, Recall@10 (R@1, R@5, R@10), Median Rank (MedR), and Mean Rank (MeanR), where higher recall and lower ranking values indicate better performance. 
\textbf{(2) Continual Learning Metric.} To quantify the model's ability to retain previously learned knowledge while learning new tasks, we adopt the Backward Forgetting (BWF), which measures the average performance drop on earlier tasks. 
Specifically, when the model is trained up to task $k$, the BWF is defined as: $\mathrm{BWF}_k = \frac{1}{k-1} \sum_{i=1}^{k-1} (R_{i,i} - R_{k,i}),$ where $R_{i,i}$ denotes the Recall@1 score on task $i$ immediately after learning task $i$, and $R_{k,i}$ represents its Recall@1 score after completing task $k$. A larger BWF implies more severe forgetting. 

\subsection{Comparison with State-of-the-Art Methods (RQ1)}
 We conduct comparison experiments on MSRVTT and ACTNET, where incremental categories are divided equally into \textit{K} (10 or 20) tasks. 

{\noindent\normalsize \bf  Evaluation on MSRVTT.} As shown in Table~\ref{talbe_m} and Fig.~\ref{figs:zxt} (a)-(b), StructAlign obtains advantages over the state-of-the-art CTVR methods and comparable performance compared to the Upper Bound: (1) it achieves the lowest Backward Forgetting and Mean Rank for all  settings; (2) it achieves the highest Recall@5 and Recall@10 for the 10 and 20 task settings; and
(3) it achieves comparable Recall@1 and Median Rank with the state-of-the-art CTVR method. 

{\noindent\normalsize \bf  Evaluation on ACTNET.} As we can see from Table~\ref{talbe_m} and Fig.~\ref{figs:zxt} (c)-(d),  StructAlign also achieves strong overall performance across all evaluated metrics on ACTNET. 
Specifically, (1) it attains the lowest Backward Forgetting across all settings; 
(2) it achieves the highest Recall@1 across all settings; 
and (3) it yields Recall@5, Recall@10, Mean Rank, and Median Rank comparable to those of the state-of-the-art CTVR method. 

{\noindent\normalsize \bf  Parameter Efficiency vs. Performance.} As shown in  Fig.~\ref{figs:pvp}, we analyze the trade-off between retrieval performance and trainable parameters by comparing the average Recall@1 across four settings. 
Most existing methods, including CLIP4Clip, X-Pool, CLIP-ViP, LwF, VR-LwF, and ZSCL, update more than 150M parameters and achieve similar average Recall@1 scores, which fall within a narrow range of 17.5-20.5. 
Parameter-efficient methods such as MoE-Adapter and StableFusion substantially reduce the number of trainable parameters (to 59.8M and 46.8M, respectively) while maintaining competitive performance. 
Our StructAlign achieves the highest average Recall@1 with 33.9M trainable parameters, offering a more favorable balance between performance and parameter efficiency.

\subsection{Ablation Studies (RQ2)}
To investigate whether performance gains are attributable to our proposed StructAlign, we conduct comprehensive ablation studies on MSRVTT and ACTNET under different task settings. 

As reported in Table~\ref{tbl2}, we draw the following conclusions:
(1) introducing the cross-modal relation preserving loss $\mathcal{L}_{CRP}$ consistently improves retrieval performance over the baseline framework on both datasets, indicating that preserving cross-modal similarity relations effectively suppresses intra-modal feature drift during continual learning;
(2) incorporating the cross-modal ETF alignment module $CETF$ further boosts performance, demonstrating that enforcing a simplex ETF geometric prior effectively mitigates non-cooperative feature drift and modality misalignment across incremental tasks;
(3) when $\mathcal{L}_{CRP}$ and $CETF$ are jointly applied, StructAlign achieves the best overall performance, which suggests that the two components are complementary and mutually reinforcing in alleviating catastrophic forgetting in CTVR. 

\subsection{Hyper-parameter Studies (RQ3)}
We conduct ablation studies on the hyper-parameters $\lambda_1$ and $\lambda_2$. As shown in Fig.~\ref{fig:ab}, for $\lambda_1$, a relatively small value of 0.1 achieves the best overall performance in both MSRVTT-10 and ACTNET-10, while higher values of $\lambda_1$ lead to slight decreases in Recall metrics, suggesting that overly strong regularization may hinder effective feature learning. 
For $\lambda_2$, moderate values around 10 yield the highest overall performance, whereas small and large values result in sub-optimal outcomes. Based on these observations, we recommend the optimal hyper-parameter setting of $\lambda_1 = 0.1$ and $\lambda_2 = 10$.

\subsection{In-depth Analysis (RQ4)} 
To better understand how the proposed method organizes the feature space, we analyze both inter-category and intra-category geometric structures. 

{\noindent\normalsize \bf  Inter-Category Analysis.}
Fig.~\ref{fig:cf} visualizes the pairwise similarity matrices of the ideal ETF prototypes and the learned text and video prototypes.
As a geometric reference, the ideal ETF exhibits uniform negative off-diagonal similarities of $-\dfrac{1}{C-1}$, corresponding to equal angular separation among categories. 
Both text and video prototypes show approximately uniform inter-category similarities and closely follow this equiangular pattern. 
This observation suggests that the learned representations exhibit well-separated category-level geometry under the proposed cross-modal ETF alignment loss. 
Moreover, the similarity patterns of text and video prototypes are  consistent, suggesting that the two modalities share a similar category-level geometric organization. 

{\noindent\normalsize \bf  Intra-Category Analysis.} 
As shown in Fig.~\ref{figs:micd}, we measure the concentration of intra-category feature distributions using the Mean Intra-Category Dispersion (MICD):
\begin{equation}
\text{MICD} = \frac{1}{C} \sum_{c=1}^{C} 
\frac{1}{|\mathcal{X}_c| \, (|\mathcal{X}_c| - 1)} 
\sum_{\substack{\mathbf{x}_i, \mathbf{x}_j \in \mathcal{X}_c \\ i \neq j}} 
\Big( 1 - \frac{\mathbf{x}_i \cdot \mathbf{x}_j}{\|\mathbf{x}_i\| \, \|\mathbf{x}_j\|} \Big),
\end{equation} 
where $C$ is the number of categories, $|\mathcal{X}_c|$ is the number of samples in category $c$, and $\mathbf{x}_i$ denotes the feature vector of the $i$-th sample. 
During training, MICD decreases from 0.3603 to 0.3369, indicating that features progressively concentrate around their category prototypes. 
Meanwhile, MICD remains strictly positive, confirming that features do not collapse and retain meaningful intra-category variability. 
This property is critical for text-to-video retrieval, as it preserves diverse instance-level semantics within each category, enabling fine-grained alignment between texts and videos. 

\section{Conclusion And Future Work}
In this work, we propose StructAlign, a structured cross-modal alignment method for the CTVR task. 
Specifically, StructAlign relies on a simplex ETF geometry to impose a structured category-level organization in the shared feature space, which helps stabilize cross-modal alignment across tasks. 
Meanwhile, a cross-modal relation preserving loss complements this geometry by regulating intra-modal feature updates through cross-modal relational constraints. 
Extensive experiments on benchmark datasets demonstrate that StructAlign effectively alleviates catastrophic forgetting and achieves competitive performance compared to state-of-the-art methods in CTVR.  
A promising direction for future work is to extend the proposed geometric and relational constraints to broader multimodal continual learning scenarios, including more open and flexible retrieval settings.  

{\small
\bibliographystyle{ieee_fullname}
\input{main.bbl}

}

\end{document}